\documentclass[12pt,a4paper]{amsart}
\usepackage{float} % 200122
\usepackage{tikz}
\usetikzlibrary{matrix,positioning,calc}
\usetikzlibrary{shapes,arrows}

\usepackage{comment}

\newtheorem{remark}{Remark} 
\newtheorem{definition}{Definition}  
\newtheorem{Exe}{Example}

\begin{document}  
 
\title[Extreme compression of grayscale images]{Extreme compression of grayscale images}

\author{Franklin Mendivil}
\email{franklin.mendivil@acadiau.ca}
 \address{Department of Mathematics and Statistics,
Acadia University,
Wolfville, NS B0P 1X0, Canada} 

\author{ \"{O}rjan Stenflo}
\email{stenflo@math.uu.se}

 \address{Department of Mathematics,
Uppsala University,
751 06 Uppsala,
Sweden
}

\today
\begin{abstract}
  Given an grayscale digital image, and a positive integer $n$, how well can we store the image at a compression ratio of $n:1$?

  In this paper we address the above question in extreme cases when $n>>50$  using  ``$\mathbf{V}$-variable image compression''.
\end{abstract}

\maketitle

\section{Introduction}

A digital $j \times k$ pixels image consists of $j \cdot k$ pixels where each pixel is assigned a ``colour'', i.e.\ an element of some finite set,
 $C$, e.g.\ $C=\{0,1,2,\ldots ,255\}^l$, where  $l=1$
 for (8 bit) grayscale images or $l=3$ for (24 bit RGB) colour images.\\
Image compression methods can be of two types:  Lossless image compression methods preserves all image data, while lossy methods removes some data from the original file and saves an approximate image with reduced file size.

One common lossless format on the internet, supported by most modern web-browsers,  is
PNG (Portable Network Graphics). The PNG format is used in particular for images with sharp colour contrast like text and line art, and is considered to be a suitable format for storing images to be edited.

One disadvantage with lossless formats is that the file sizes are often very large when compared with lossy formats.
Lossy compression of colour images is often obtained by reducing the colour space, or by 
chroma subsampling using the fact that the human eye perceives changes in brightness more sharply than changes in colour. 

The most common lossy compression format, especially for photographic images, is JPEG (Joint Photographic Experts Group).
JPEG usually gives small file sizes, but one artifact with JPEG is apparent ``halos''  in parts of the image with sharp colour contrasts, reflecting the fact that the JPEG method belongs to the realm of Fourier methods.
%using chroma subsampling.
Similar ``halo'' features are also apparent in the better, but more complicated and therefore less widespread, wavelet based JPEG 2000 method.

% satellite images

%Lossless:
%BMP: Run-length encoding \\
%Suitable for simple graphics files with long runs of identical data \\
%PNG, ZIP, gip: Deflate (non-patented) (LZ77 algorithm and Huffman coding) \\
%GIF, TIFF: LZW (small data analysis) \\

%Lossy:
%GIF, PNG: reduce color space 
%JPEG: Transform encoding (DCT), Chroma subsampling (human eye perceives changes% in brighness more sharply than changes in colour. Avareges chroma (colour) mai%ntains luma (brightness))

%Rare:
%TIFF (extremely wide,non-supported), PCX;TGA, etc.

The mathematical  notion of $V$-variability was introduced by Barnsley, Hutchinson and Stenflo in \cite{Barnsleyetal05}.
%,  \cite{Barnsleyetal08}, \cite{Barnsleyetal12}.
Intuitively, a $V$-variable set is built up by at most $V$ different smaller sets at any given level of magnification.
Motivated by the fact that parts of an image often resemble other parts of the image, and the existing ``fractal compression method'' building on a more limited class of
 sets \cite{BarnsleySloan88},
it was suggested  in \cite{Barnsleyetal05} that $V$-variability could be used for image compression.

One solution to the problem of how the notion of $V$-variability can be used for image compression was presented in Mendivil and Stenflo
\cite{MendivilStenflo15} in terms of an algorithm for lossy image compression where, for  a given digital image, the algorithm generates a
``$V$-variable digital image'' resembling the given image despite requiring substantially less storage space.

In the present paper we address the question of finding an ``optimal''
$V$-variable digital image resembling a target image given a certain maximally allowed storage space.

\section{ $\mathbf{V}$-variable image compression}
For clarity and simplicity of our description we will assume here that the image is of size $2^m \times 2^m$ pixels for some $m$.
We refer to Section \ref{nsq} for the case of  non-square images.

In order to specify what we mean by a ``$V$-variable digital image''
(and later a ``$\mathbf{V}$-variable digital image'') we need to introduce some definitions.

\begin{definition} 
Let $m \geq 0$ be some integer, and consider a $2^m \times 2^m$ pixels digital image.  
For any $0 \leq n \leq m$, we may divide the given
 image  into
$4^n$ nonoverlapping image pieces of size $2^{m-n} \times 2^{m-n}$.
We call these pieces the image pieces of  level $n$.
\end{definition}

We can now define what we mean by a $V$-variable image:

\begin{definition}
Let $V \geq 1$ be a fixed positive integer. We say that a digital image is \emph{$V$-variable} 
%(w.r.t. a given sequence of refining  partitions)
if, for any  $1 \leq n \leq m$, the image has at most $V$ distinct image pieces
 of level $n$.
\end{definition}
 
\begin{Exe} \label{4varEx} 
The $4$-variable $512 \times 512=2^{9} \times 2^{9}$ pixels digital  grayscale image
\begin{center}
\includegraphics[width=1.5in]{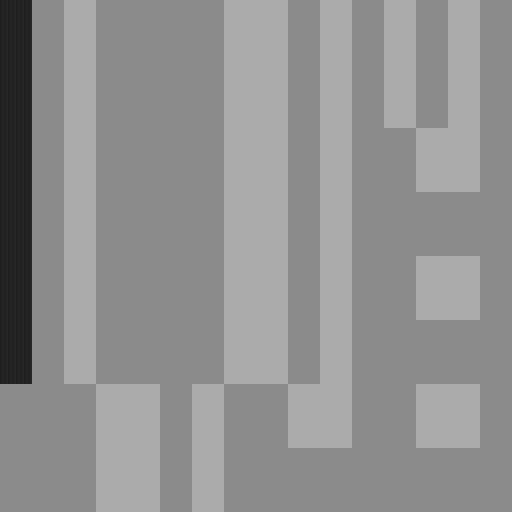} \\ 
\end{center}
can be built up by $4^n$ image pieces of size 
$2^{9-n} \times 2^{9-n}$ of (at most) $4$ distinct types, for any
 $n=0,1,2,\ldots ,9$.
The appearance of these image pieces depends on $n$.
If e.g.\  $n=2$ then the $16$  image pieces of size $128 \times 128$ pixels are of the $4$ types
\begin{center} 
    \resizebox{!}{14mm}{\includegraphics{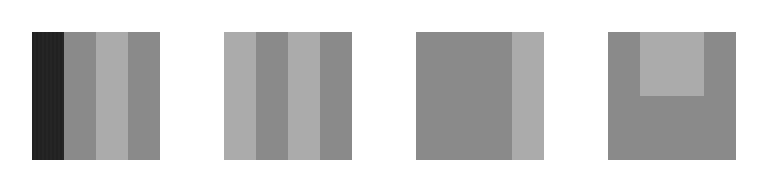}} \label{Enya4_4_montage}
\end{center}
%\hspace{23mm}\parbox{12cm}{{\em  The 16 blocks at level 2 of the Enya4 imag%eare of 4 possible types}}
and if $n=3$ then  the $64$  image pieces of size $64 \times 64$ pixels
are of the $4$ types
\begin{center} 
   \resizebox{!}{7mm}{\includegraphics{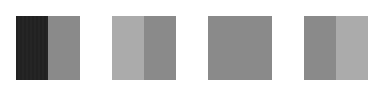}} \label{Enya4_8_montage}
\end{center}

\noindent
By looking at the image, and its image pieces,  we see that we can, recursively, describe 
the $4$-variable image using $4$ images of smaller and smaller size, i.e.\ recursively describe more and more levels of the $V$-variable structure of the image:  

\begin{center} 
   \resizebox{!}{100mm}{\includegraphics{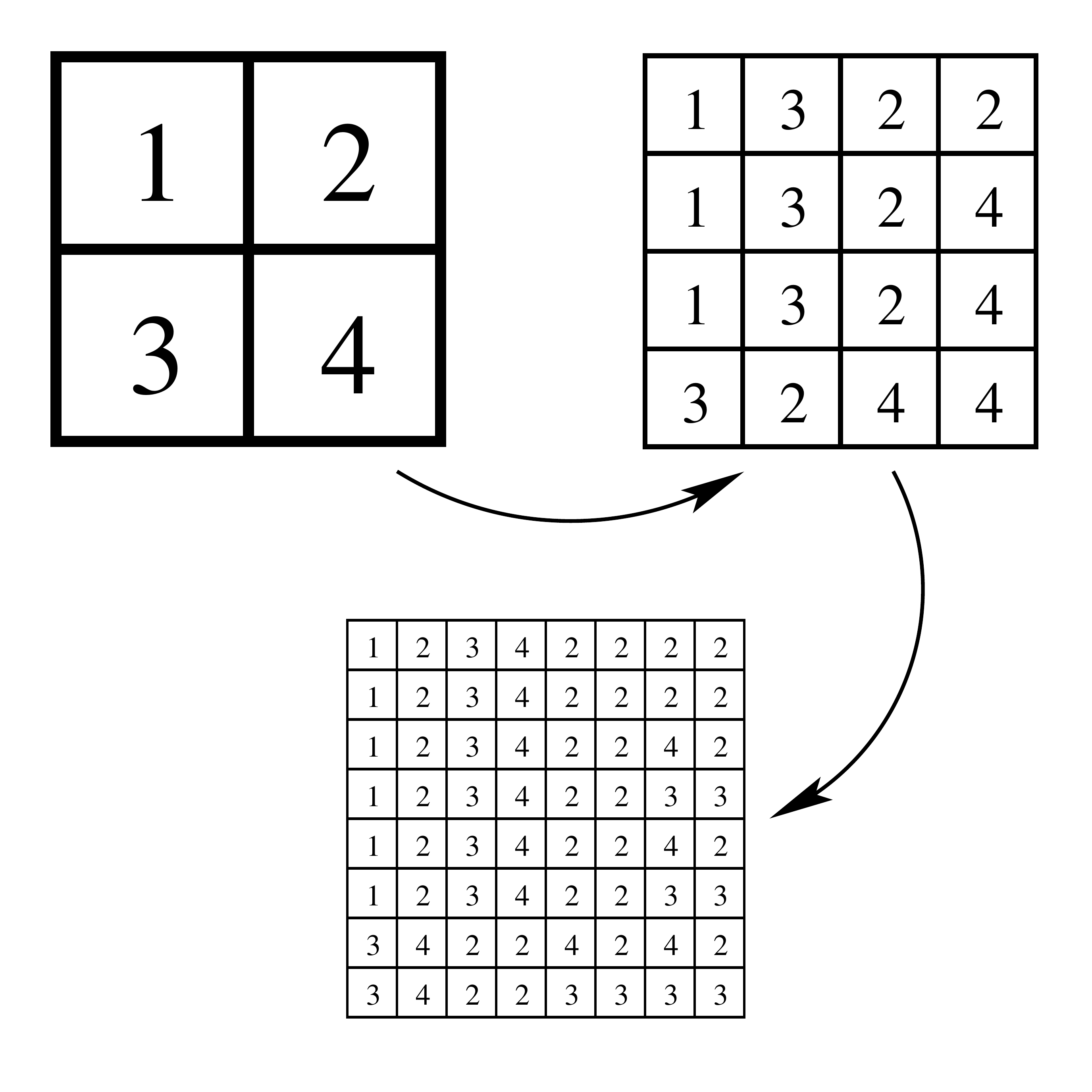}} 
\end{center}

At each stage, we replace an image piece of the current level with four image pieces from the next level.   There are (at most) $V=4$ types of
image pieces at each level and the substitution is done according to the type.  For example, we see that in the second stage (illustrated in the figure above),  all image pieces of type $3$ are replaced by the same thing (image pieces at next level with types $3,4,3,4$).  The first two (non-trivial) steps as shown above can be visually described by the substitutions:

\begin{center} 
   \resizebox{!}{60mm}{\includegraphics{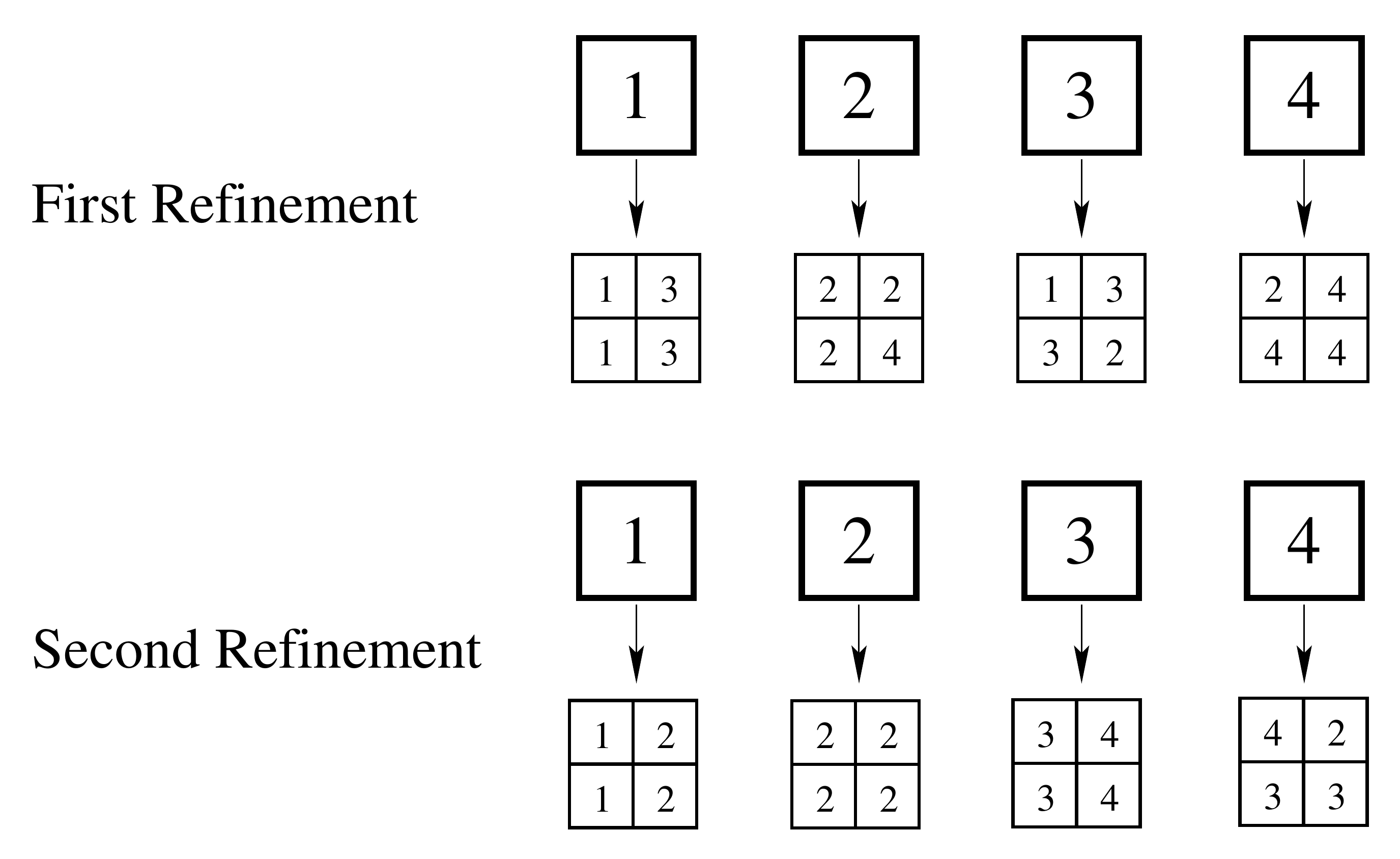}} 
\end{center}
%\hspace{23mm}\parbox{12cm}{{\em  The 64 image pieces at level 3 of the Enya4 imag%eare of 4 possible types}}
%(An alternative way of visualising the matrix $S$ describing the skeleton tree.)

Each image piece of level 9 is a one pixel image. For such pieces we associate the corresponding colour value.

In this $4$-variable example the image is constructed using only the $4$ colour values
$33$ and $37$ (corresponding to the almost black pixels) $138$ (corresponding to the dark gray pixels) and $171$ (corresponding to the light gray pixels). 

See \cite{MendivilStenflo15} and Section \ref{subsec:reconstruction} for further details and the specifics on how to reconstruct the image from its code (the substitution rules).
\end{Exe}

\subsection{Description of our $V$-variable image compression method}  \label{method}

The goal of our method is to approximate a given digital image with a $V$-variable digital
image.

We start with positive integers $V_1, V_2, \ldots, V_m$ which restrict the number of % allowed ``types'',
distinct image pieces, 
of each level $n$, $n=1,2,\ldots, m$.
Note that by our choice of partitions at each of the ``levels'', there are at most four times as many distinct image pieces
of level $n+1$ as there were at level $n$.
Thus any value of $V_n$ which is at least $\min(4^n,4V_{n-1})$ provides no constraint that is not already forced by $V_1,\ldots,V_{n-1}$.

Let $V=\max(V_1,\ldots,V_m)$.
%and let $n_0$ be the smallest integer such that
%$V_{n_0} < 4^{n_0}$.
We can find a $V$-variable approximation of the given image, with $V_1$,\ldots,$V_m$ being restrictions on the maximum number of allowed distinct image pieces of each level respectively, by using the following algorithm:\\
%given in Figure \ref{fig:algorithm}.\\
\newpage

\begin{figure}[hp]   
\begin{center}
% Define block styles
\tikzstyle{decision} = [diamond, draw, fill=blue!10, 
text width=8em, node distance=6cm, inner sep=0pt]
\tikzstyle{vardecision} = [diamond, draw, fill=blue!10, 
    text width=8em, text badly centered, node distance=10cm, inner sep=0pt]
\tikzstyle{block} = [rectangle, draw, fill=blue!10, 
text width=12em, rounded corners, minimum height=4em]
\tikzstyle{vblock} = [rectangle, draw, fill=blue!10, 
    text width=10em, node distance=5.3cm, rounded corners, minimum height=4em]
\tikzstyle{varblock} = [rectangle, draw, fill=blue!10, 
    text width=20em, node distance=10cm, rounded corners, minimum height=4em]
\tikzstyle{line} = [draw, -latex']
\tikzstyle{cloud} = [draw, ellipse,fill=red!20, node distance=1cm,
    minimum height=2em]
    
\begin{tikzpicture}[thick, scale=0.6, transform canvas={scale=0.6}, node distance = 3cm, auto]
    % Place nodes
    \node [block] (Nr1) {1.Get image and choose $V_1,\ldots,V_m$};
   % \node [cloud, left of=init] (expert) {expert};
    \node [block, below of=Nr1] (Nr2) {2.Extract all image pieces of level $j=n_0$ where $n_0$ is the smallest integer such that $V_{n_0}<4^{n_0}$};
    \node [block, below of=Nr2] (Nr3) {3.Classify all image pieces into $V_j$ clusters and identify each cluster with a cluster representative image};
    \node [block, below of=Nr3] (Nr4) {4.Store classification of image pieces};
     \node [block, below of=Nr4] (Nr5) {5.Divide each image representative into 4 image pieces, and consider the $n$ image pieces created};
     \node [vardecision,  right of=Nr3] (Nr7) {\ \ \ Is $n >V_{j}$?};
      \node [vblock,  right of=Nr4] (Nr8) {8. Regard the image pieces as image representatives};
     \node [decision, below of=Nr5] (Nr10) {Are image pieces pixels?};
      \node [vblock, below of=Nr10] (Nr6) {6.Store colour of image pieces};
      \node [varblock, right of=Nr10] (Nr9) {7.\hspace{2cm} \ \ \ \ \ \ \ $j:=j+1$};
      \path [line] (Nr7) -- node [near start] {yes} (Nr3);
      \path [line] (Nr7) -- node [near start] {no} (Nr8);
      \path [line] (Nr10) -- node [near start] {no} (Nr9);
       \path [line] (Nr10) -- node [near start] {yes} (Nr6);

    % \path [line] (Nr5) -- node [near start] {no} (Nr8);
   % \node [block, below of=decide, node distance=3cm] (stop) {stop};
    % Draw edges
     \path [line] (Nr1) -- (Nr2);
     \path [line] (Nr2) -- (Nr3);
    \path [line] (Nr3) -- (Nr4);
    \path [line] (Nr4) -- (Nr5);
    \path [line] (Nr4) -- (Nr5);
    \path [line] (Nr5) -- (Nr10);
     \path [line] (Nr9) -- (Nr7);
 \path [line] (Nr8) --  (Nr5);
      
%\path [line] (Nr6) -| node [near Nr3] {yes};
   % \path [line] (update) |- (identify);
   % \path [line] (decide) -- node {no}(stop);
   % \path [line,dashed] (Nr6) -- (Nr3);
   % \path [line,dashed] (system) -- (init);
   % \path [line,dashed] (system) |- (evaluate);
\end{tikzpicture}

% \caption{Outline of $V$-variable compression algorithm.}  \label{fig:algorithm}
\end{center}
\end{figure} 

\newpage

In  \cite{MendivilStenflo15}, we formulated and applied the above algorithm for various choices of $V$ and images,
where $m=9$ and $V=V_1=\ldots=V_m$ (and so we only considered images which had the same ``maximal variability'' at each level).

The above algorithm with non-constant $V_n$ generates a $V$-variable image for $V=\max(V_1,\ldots,V_m)$ with less memory requirements than the $V=V_1=\ldots=V_m$ case considered in \cite{MendivilStenflo15}, but the image quality is also decreased.  
The art is then to try to find the optimal $m$-tuple of positive integers  ${\mathbf{V}}=(V_1,\ldots,V_m)$ with respect to targeted image quality and memory requirements.

We will present computer experiments related to this question for some given test images
below. 

See Section \ref{general} for a discussion of some   generalizations of the algorithm.

\begin{definition} Let  ${\mathbf{V}}=(V_1,\ldots,V_m)$, and $V=\max(V_1,\ldots,V_m)$, where $V_1,\ldots,V_m$ is a  given sequence of positive integers.
  We say that an \linebreak[4] $2^m \times 2^m$ pixels digital image is
  $\mathbf{V}$-variable if it is $V$-variable with maximally $V_i$
distinct image pieces of level $i$, for any  $i=1,\ldots,m$.
\end{definition}
\subsection{Storage requirements} \label{StorReq}
The $m$-tuple of positive integers \linebreak[4]
${\mathbf{V}}=(V_1,\ldots,V_m)$ , chosen by the user in step 1 in the algorithm above,
corresponds to the quality of the image approximation. 
Larger values of the components corresponds to higher image quality of the constructed $\textbf{V}$-variable image at the price of a larger file size. We may without loss of generality assume that
$V_i \leq \min(4^i,4V_{i-1}) $ for all $1 \leq i \leq m$ (with the convention $V_0=1$):

% The first
%Every time step $4$ is reached, $4^{n_0}$ numbers from the range
%\linebreak[4] $\{1,\ldots, V_{n_0}\}$ have to be stored.
%After this,
Each execution of step $4$ requires a storing of $4 V_{j-1} $ numbers from the range  $\{1,\ldots, V_{j}\}$ 
for any $j$, $n_0 \leq j \leq m-1$.
% with $V_{j}<4 V_{j-1}$.
The image pieces extracted in step $2$  are of size $4^{m-n_0}$ pixels, and each time step $5$ is reached the new image pieces are $4$ times smaller than the ones from the previous execution of step 5.
Thus step $4$ will be revisited at most $m-n_0-1$ times.

In step $6$ we store the  colour, i.e.\  an element in $C$, 
 for each of the $4V_{m-1}$ ``one pixel'' image representatives.
 As an example, for the 8 bit grayscale images considered in this paper
 % in Figure \ref{fig:testimages}, $C=\{0,1,\ldots, 255\}$, and
 we need to store $4 V_{m-1}$ numbers in $\{ 0, \ldots, 255 \}$, and we may
 thus, in order to optimize the image quality for a given storage space, without loss of generality assume that $V_m=256$ since the storage does not depend on $V_m$. 
 
 Thus, our $V$-variable method of storing requires in total  
%\[
%   4^{n_0} \lceil{\log_2( V_{n_0})}\rceil/8 = 2^{n_0-3}  \lceil{\log_2( V_{n_0}%)}\rceil \ \rm{bytes}
%\] 
%for  level $n_0$ plus
\linebreak[4] $4V_{j-1} \lceil{\log_2( V_{j}}\rceil/8 $ bytes for each level $n_0 \leq j \leq m-1$,  where $4V_{j-1} > V_{j}$,
 plus $4V_{m-1}$ bytes for  level $m$ in the 8 bit grayscale images case.

\begin{remark}
   We will here only consider the case when $m=9$ and $4V_k \geq V_{k+1}$ for all $k$.
   In order  to avoid a waste of memory we will also only consider cases when
   $V_k=2^{j_k}$ for some $j_k$ for all $k$, where $V_{n_0-1}=4^{n_0-1}$  giving a total storage (in the case $m=9$) of
\[
  \text{Storage}=
\sum_{k \in I} 4V_{k-1} \lceil{\log_2( V_{k}}\rceil/8+ 4V_{8}=
  \sum_{k \in I} j_{k} 2^{j_{k-1}-1}+ 2^{j_8+2} \ \ \  \text{ bytes},
 \]
where $I= \{ n_0 \leq k \leq 8: V_k < 4V_{k-1} \}$. Each $ k $ with $V_{k}=4V_{k-1}$ gives no contribution to the sum so 
 for instance if $V_4=4^4=256$, $V_5=2^8=256 < 4^5$, $V_6=2^5=32$,
 $V_7=4V_6=128$, $V_8=2^6=64$ then $n_0=5$, $j_4=8$, $j_5=8$, $j_6=5$, $j_7=7$,  and $j_8=6$ and thus
\[
  \text{Storage}= 8 \cdot \underbrace{2^{8-1}}_{128}  + 5 \cdot \underbrace{2^{ 8-1}}_{128} + 0 + 6 \cdot \underbrace{2^{ 7-1}}_{64}   + \underbrace{ 2^{ 6+2}}_{256}
  =2304 \ \ \  \text{ bytes},
\]
where the zero term occur since $4V_6=V_7$, and thus $V_7$ causes no new restrictions not already caused by $V_1$,\ldots,$V_6$.
%If, on the other hand, $V_4=256=4^4$, $V_5=V_6=V_7=V_8=128 \leq 1023$, then  $n%_0=5$,$j_5=j_6=j_7=j_8=7$ and thus
%\[
%   \text{Storage}= 256 \cdot 7/2  + 7 \cdot 64 + 7 \cdot 64 + 7 \cdot 64 + 512=%2880 \ \ \  \text{ bytes}.
%\]
\end{remark}

\subsection{Reconstruction of an image from its code}
\label{subsec:reconstruction}

The process of reconstructing an image  from its $V$-variable code is quick.\\

Any pixel sits within some fixed image piece of level $n$ for all $0 \leq n \leq m-1$.
An image piece of level $n$, $1 \leq n \leq m$, is situated within a 4 times bigger image piece from level $n-1$ and can be in 4 possible positions in that bigger square:\\[5pt]
\hbox{} \qquad 1) upper left, or  \quad
2) upper right, or  \quad
3) lower left, or \quad
4) lower right.\\[5pt]
This enables a simple addressing structure, where a pixel can be described by an address $a_1 a_2...a_m$, where $a_j \in \{1,2,3,4\}$ specifies the relative position of the square on level $j$ in the square on level $j-1$ for the given pixel.\\
 
We can now  find the colour of a pixel by recursively using the $V$-variable code in the following way.
If a pixel with address $a_1 a_2...a_m$  sits within a square with $V$-variable type $j$ of level $n$, $0 \leq n \leq m-1$, then the $V$-variable type of the square on level $n+1$ where the pixel sits can be seen by looking at the stored type of the square in position $a_{n+1}$ within the square of type $j$ on level $n$. The colour of the pixel is given by the stored colour of the square in position $a_m$ of the type of the square at level $m-1$.

\subsection{Automating the $V$-variable image compression algorithm} \label{auto}
%We implementated our compression method in Matlab, and  used Matlab's built-in %command
%{\tt kmeans} for the crucial third step in our algorithm.  

The main tool needed in order to automate the algorithm above is a way to classify $n$ images into $k$ clusters. (step 3).
Such a clustering can be done in many different ways.  See Section \ref{clusterm} for a discussion of some basic clustering techniques.
For simplicity, in our implementation % in Examples \ref{ex1} and \ref{ex2}
of the compression algorithm above we  used Matlab's built-in command
{\tt kmeans} for this step. 
 The K-means algorithm is a popular and basic clustering algorithm which  finds clusters and cluster representatives for a set of vectors by iteratively minimising
the sum of the squares of the ``within cluster''  distances (the distances from each of the vectors to the closest cluster representative). % \cite{HTF}.
  We  treat the sub-images as vectors and use the standard Euclidean distance to measure similarity.  We also use random initialisation of the cluster representatives.

\subsection{Finding the optimal $\mathbf{V}$ for a given memory requirement}
For a given image and tuple $\mathbf{V}=(V_1,\ldots,V_m)$ we may use the algorithm described in
Section \ref{method} to generate a $\mathbf{V}$-variable approximating image of the given image and in  Section \ref{StorReq} we explained how to calculate an upper bound for the memory requirement for a $\mathbf{V}$-variable image.

A natural problem that arises is to find the optimal $\mathbf{V}$ for a given memory requirement with respect to image quality.
The optimal choice may depend on the image, but in general simulations show that a ``good'' $\mathbf{V}$ is often ``good'' for most images.

Rather than finding the optimal choice for a a given image and memory requirement we will here give rules of thumb for choices of $\mathbf{V}$ for some given memory requirements that may be used for any image.

\subsection{Simulations}

\begin{figure}[h!]   %[hp]   
\includegraphics[width=1.6in]{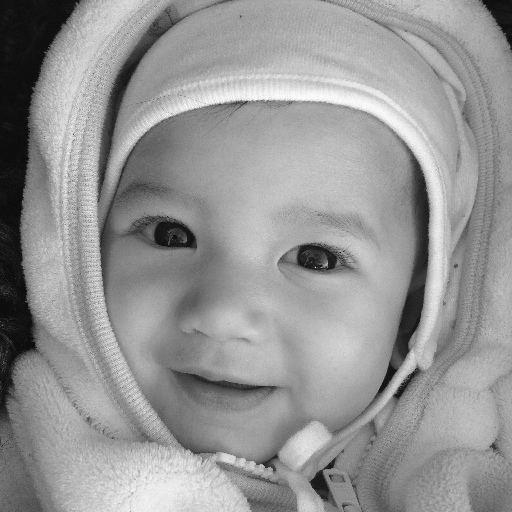} %= Enya.png
\includegraphics[width=1.6in]{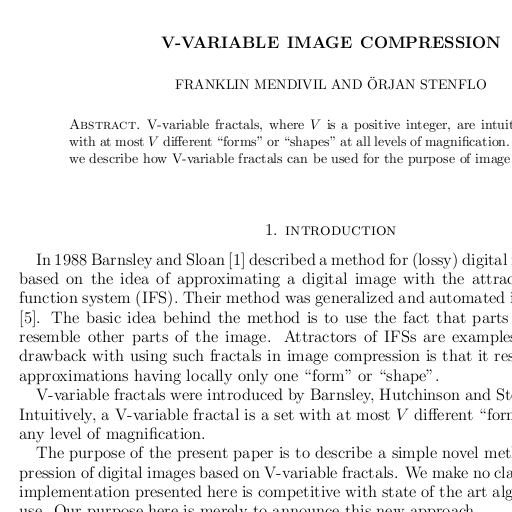} %=PaperPNG.png 
\includegraphics[width=1.6in]{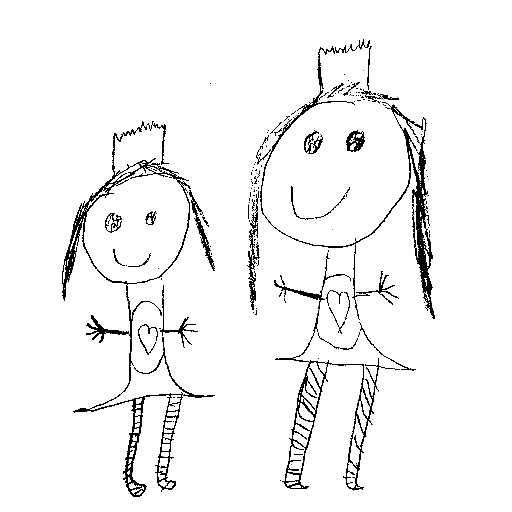} %=Art.png
\caption{Three test images: Image1, Image2, Image3} \label{fig:testimages}
\end{figure}

In this section we present results of some numerical experiments with the three images shown in Figure \ref{fig:testimages}.
The images are of size $512 \times 512$ pixels
so uncompressed they are stored with $512 \cdot 512=262144$ B (since each pixel is assigned a number in  $\{0,1, \ldots, 255\}$ 
and thus requires one byte of storage). Thus approximations stored e.g.\ with a file size of 2621 B correspond to a 100:1 compression ratio.
As can be seen, the three images represent a  wide range of image types.

Image1 and Image2 have earlier been discussed in \cite{MendivilStenflo15}.\\

\begin{figure}[h!]
  \includegraphics[width=4.0in]{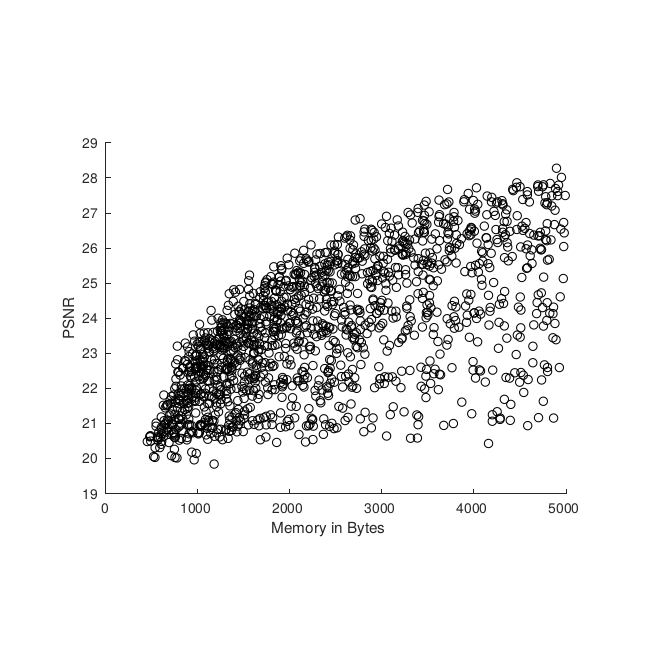}
  \caption{Scatterplot of filesize versus PSNR for Image1}
  \label{fig:scatterenya} 
\end{figure}

\noindent
We will here use the Peak signal-to-noise ratio (PSNR) for measuring quality of compressed images. 
%PSNR=10*log_{10}( 255^2/MSE )
Figure \ref{fig:scatterenya} shows the quality of $\mathbf{V}$-variable approximations of Image1 (PSNR) versus memory requirement for all $\mathbf{V}=(V_1,\ldots,V_9)$  requiring a memory less than 5000 B
with $V_1=4,V_2=16,V_3=64$ and $V_9=256$ and $V_4, V_5,V_6,V_7$ and $V_8$
chosen from $\{16,32,64,128,256,1024\}$.

% \label{fig:approxfigurs}

%\end{center}
%\end{figure}

The frontier in the plot gives a hint of how the optimal PSNR depends on the memory requirement.

The table in Figure \ref{thumb} gives rules of thumb for good tuples $\mathbf{V}$ for  given memory requirements of  $1000\mathrm{B},1500\mathrm{B},\ldots,5000\mathrm{B}$ respectively.
These $\mathbf{V}$:s corresponds to points close to the frontier in
Figure \ref{fig:scatterenya}. In simulations below ``ImageI-Memory'' denotes the $\mathbf{V}$-variable image obtained from the algorithm
to approximate ImageI, I $\in \{1,2,3\}$, with a maximal allowed memory  of Memory $\in \{500,1000,\ldots,5000 \}$ Bytes
using the rules of thumb for $\mathbf{V}=(V_1,\ldots,V_9)$ given by Figure
\ref{thumb}.

% We refer to Section \ref{sec3} for a comparison with other compression methods.

The image labeled ImageI:256 % and ImageI:1024
correspond to the best $256$-variable approximation w.r.t.\ PSNR
% and $1024$-variable approximations
and has
earlier been discussed in \cite{MendivilStenflo15}. 

\begin{figure}[h!]
%\begin{table}[htb] 
\begin{tabular}{ |p{23mm}||p{7mm}|p{7mm}|p{7mm}|p{7mm}|p{7mm}||p{13mm}|p{13mm}|  p{13mm}|p{13mm}|}  
  %\multicolumn{8}{|c|}{$V$-variable approximations of Image1} \\
  \hline
  Label         & $V_4$     & $V_5$   &   $V_6$  & $V_7$  & $V_8$     & Memory (B) & PSNR Image1 &  PSNR Image2 &  PSNR Image3\\
  \hline
 ImageI-500 & 16  & 16   & 16 & 16  & 64  &  480 & 20.6 & 15.1 & 13.9\\
 ImageI-1000  & 256 & 32   & 16 & 16  & 64  &  992 & 23.5 & 15.9 & 14.7 \\
ImageI-1500   & 256 & 64   & 64 & 32  & 64  &  1472 & 24.5 & 16.3 & 15.3 \\
ImageI-2000   & 128 & 512  & 32 & 32  & 64  &  1936 &   25.4 & 16.9   &   16.5\\
ImageI-2500   & 256 & 256  & 32 & 128 & 64  &  2304 & 25.9 & 17.1 & 16.8 \\
 ImageI-3000  & 256 & 256  & 64 & 128 & 128 &  2976 & 26.5 & 17.4 & 17.2 \\
  ImageI-3500  & 256 & 256 & 1024 & 16  & 64 &  3328 & 26.8 
                                                                                                 & 17.9 & 18.4 \\
   ImageI-4000              & 256 & 1024 & 64 & 128 & 64  &  3936 &27.5 & 18.6 & 17.3 \\
  ImageI-4500   & 256 & 256 & 1024 & 64  & 64 &  4544 & 27.7 & 18.5 & 20.0 \\
ImageI-5000   & 256 & 1024 & 128 & 128 & 128  &  4992 &28.1 & 19.1 & 17.9 \\
  % & 256  & 32 & 32  & 32 & 32 & 32  &   960      & 23.6      \\
  %              &  256  & 64 & 64  & 64 & 64 & 64  &   1536      &   24.5    \\
 %               &  256  & 64 & 64  & 64 & 64 & 256  &    1600     &  24.8     \\
  % &  256  & 256 & 64  & 64 & 32 & 32  &   2224      & 25.8      \\
      %          & \textcolor{red}{ 256}  & \textcolor{red}{128} &
%                                                                  \textcolor{red}{128}  & \textcolor{red}{128} &\textcolor{red}{128} & \textcolor{red}{128}  &  \textcolor{red}{ 2688 }    & \textcolor{red}{ 25.9 }    \\
 %               &  256  & 256 & 64  & 64 & 64 & 32  &   2336      & 26.0      \\
 % &  256  & 256 & 128  & 64 & 32 & 32  &   2544      & 26.0      \\
% Image1:256-64   & \textcolor{blue}{ 256}  & \textcolor{blue}{256} &%
          %                                            \textcolor{blue}{64}  & \textcolor{blue}{64} &\textcolor{blue}{64} & \textcolor{blue}{64}  &  \textcolor{blue}{ 2368 }    & \textcolor{blue}{ 26.2 }    \\
 %               &  256  & 256 & 128  & 64 & 64 & 32  &   2656      & 26.2      \\
  % &  256  & 256 & 128  & 128 & 128 & 128  &    3264     &  26.6     \\
   % &  256  & 256 & 256  & 64 & 64 & 32  &   3168      & 26.7      \\
   %  &  256  & 256 & 256  & 256 & 64 & 32  &   4000      & 26.9      \\
   %             &  256  & 256 & 256  & 256 & 256 & 32  &   4736      & 27.2      \\
         ImageI:256          & 256 & 256  & 256 & 256  & 256  &   5120      & 27.4  & 18.2 & 19.2    \\
  \hline
  \end{tabular}
  % \end{table}
  \caption{Suggested values of $\mathbf{V}=(V_1,\ldots,V_9)$ gives rules of thumb
    for a given upper bound memory requirement. Here $V_1=4,V_2=16,V_3=64$, and $V_9=256$.}
  \label{thumb} 
\end{figure}

For most $\textbf{V}$-variable images we may store the code in a much more efficient way than described above, so the above memory requirements should be regarded as upper bounds.

 Let $i$ be a non-negative integer. By regarding all ``almost constant'' blocks with pixel values varying less than or equal to $i$, as constant blocks in each step of the construction of the approximating image,
%and decrease the colour space to $2^c$ colours, where $c$ is a positive integer% less than or equal to $8$,
 we can reduce the storage further at a small price in image quality  if $i$ is small.
 To illustrate this method, let us look at  $\textbf{V}=(V_1,\ldots,V_9)=(4,16,64,256,256,32,128,64,256)$ (our rule of thumb \textbf{V} for a storage requirement less than 2500 B).
Below we have for each $i \in \{0,15,30 \}$ simulated a $\textbf{V}$-variable approximating image of Image1, where we, in each step of the construction, regard blocks with pixel values varying less than or equal to $i$ as constant.
We refer to the corresponding image as ``Image1-2500-i''.
 
The following tabular shows the proportion of constant substitutions at each level:\\

\begin{small}
  
\begin{tabular}{ |p{12mm}||p{2cm}|p{23mm}|p{23mm}|p{23mm}| }
 \hline
  \multicolumn{5}{|l|}{Image1-2500-i % ($V_1=V_2=V_3=V_4=V_5=256,V_6=32,V_7=128,V_8=64,V_9=256$)
  } \\
 \hline
  Level & Block size  &  Proportion of constant substitutions ($i=0$) &
                                                                         Proportion of constant substitutions ($i=15$) &  Proportion of constant substitutions ($i=30$)
  \\
  \hline
 $ 4$   &  $32 \times 32$  & 4/256     & 6/256 & 8/256   \\
 $5$   &  $16 \times 16$ & 24/256    & 26/256 & 36/256   \\
 $6$        &  $ 8 \times 8 $ & 0    & 0  & 0  \\
 $7$ & $ 4 \times 4 $      &  35/128 & 45/128  & 75/128   \\
  $8$    & $2 \times 2$ & 2/64 & 32/64  &  51/64 \\
  \hline
 % Storage &       & 2162 B   & 2040 B & 1891 B\\
 % \hline
\end{tabular} \\
\end{small}

Recall that if
$V_k=2^{j_k}$ for some $j_k$ for all $k$, where $V_{n_0-1}=4^{n_0-1}$  then the upper bound for storage where we ignored the information about constant blocks was
\[
  \text{Storage}=
  \sum_{k \in I} j_{k} 2^{j_{k-1}-1}+ 2^{j_8+2} \ \ \  \text{Bytes},
 \]
 where $I= \{ n_0 \leq k \leq 8: V_k < 4V_{k-1} \}$.

If $p_k$ denotes the proportion of constant substitutions at level $k$, $n_0 \leq k \leq 8$, then since we need only one instead of four numbers to store each such substitution,  the storage can be reduced to 
\[
  \text{Storage}=
  \sum_{k \in I} j_{k} 2^{j_{k-1}-3}(4-3p_{k-1})+ 2^{j_8}(4-3p_8) \ \ \  \text{Bytes}.
 \]

For Image1-2500-0, $V_4=4^4=256$, $V_5=2^8=256 < 4^5$, $V_6=2^5=32$,
 $V_7=4V_6=128$, $V_8=2^6=64$ i.e.\ $n_0=5$, $j_4=8$, $j_5=8$, $j_6=5$, $j_7=7$,  and $j_8=6$, and $p_4=4/256$, $p_5=24/256$, $p_7=35/128$, and $p_8=2/64$. Thus  
\begin{eqnarray*}
  \text{Storage(Image1-2500-0)}&=&  8  \cdot 32(4-3 \cdot \frac{4}{256})  +
 5 \cdot 32(4-3 \cdot \frac{24}{256}) + 0 \\ & &+
  6 \cdot 16(4-3\cdot \frac{35}{128})   + 64(4-3\cdot \frac{2}{64}) \approx 2162 \text{B}. 
\end{eqnarray*}
Similarly
\begin{eqnarray*}
  \text{Storage(Image1-2500-15)}&=&   8  \cdot 32(4-3 \cdot \frac{6}{256})  +
 5 \cdot 32(4-3 \cdot \frac{26}{256}) + 0 \\ & &+
  6 \cdot 16(4-3\cdot \frac{45}{128})   + 64(4-3\cdot \frac{32}{64})=2040 \text{B},
\end{eqnarray*}
and
\begin{eqnarray*}
  \text{Storage(Image1-2500-30)}&=&
    8  \cdot 32(4-3 \cdot \frac{8}{256})  +
 5 \cdot 32(4-3 \cdot \frac{36}{256}) + 0 \\ & & +
  6 \cdot 16(4-3\cdot \frac{75}{128})   + 64(4-3\cdot \frac{51}{64})\approx 1891 \text{B}.
\end{eqnarray*}

The below scatter plots, where we have plotted memory requirement versus PSNR for  $\textbf{V}$-variable approximations of our 3 test images, illustrates the interesting phenomena
that we  get approximations with both  higher PSNR and lower storage space if we regard all blocks varying less than a given threshold
as being constant (the average of the grayscale values).
The threshold depends on the given memory requirement and decreases with increased memory requirements.

  \begin{figure}[H] 
\begin{center}
\begin{tabular}{cc}
\includegraphics[width=3in]{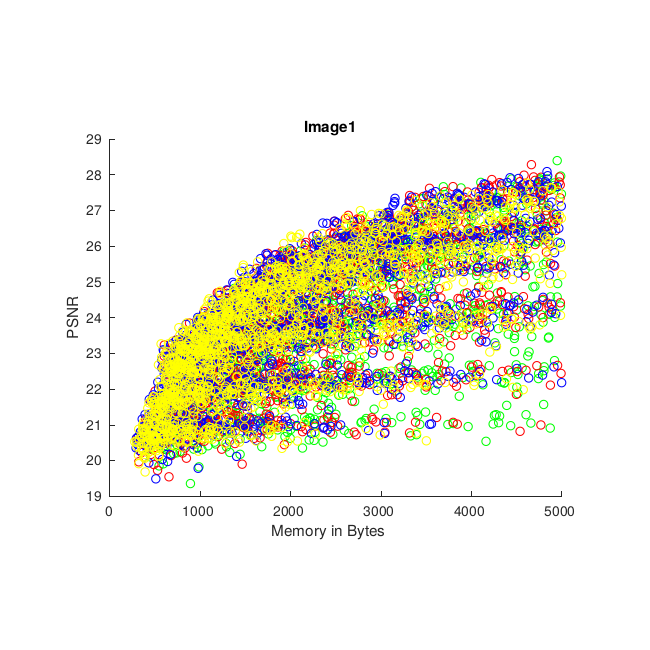}  &
                                                \includegraphics[width=3in]{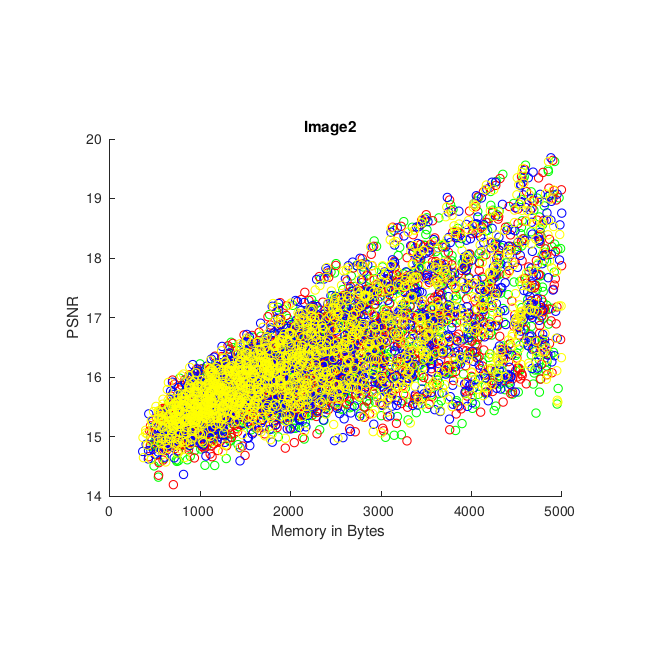} \\
\includegraphics[width=3in]{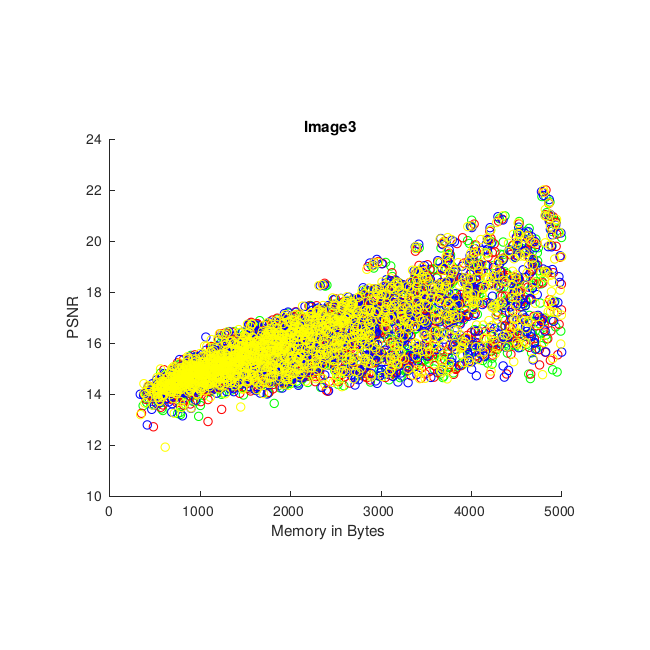}  &
\end{tabular}
%\vspace{-2cm}
\caption{Scatterplots of memory versus PSNR for Image1 (upper left plot), Image2 (upper right plot) and Image3 (lower left plot).
  Points where blocks varying less than a given threshold have been regarded as constant
have been marked with different colours, here 
green (0), red (15), blue (30) and yellow (45), where the threshold is given within the brackets.
The scatterplots indicates that a threshold of 30 is optimal with respect
to PSNR and memory, for  compression ratios around 150:1,
and the optimal threshold tends to decrease with file size.
  %3000 B (PSNR 17.5), and 4000 B (PSNR 17.9), 5000 B (PSNR 18.1), 6000 B (PSNR 18.1).  
}
\end{center}
 %\begin{tabular}{c}
 % \includegraphics[width=6in]{Scatter256topPaperPNG}
 % \end{tabular}
\end{figure}

%We can thus find a 256-variable approximating image with PSNR slighlty above PS%NR 25 at compression ratio 150:1 and we obtain a PSNR around 26.5 at compressio%n ratio 100:1.

%\begin{tabular}{c}
 
%\end{tabular}

\begin{figure}[h!] %[hp]  
%\begin{center}
\begin{tabular}{cc}
\includegraphics[width=2in]{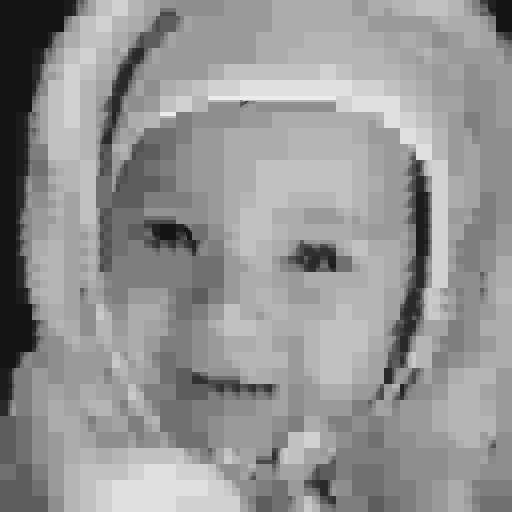} &
\includegraphics[width=2in]{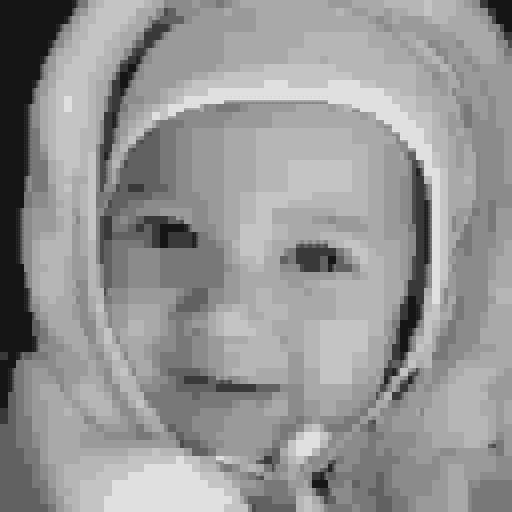} \\ 
\includegraphics[width=2in]{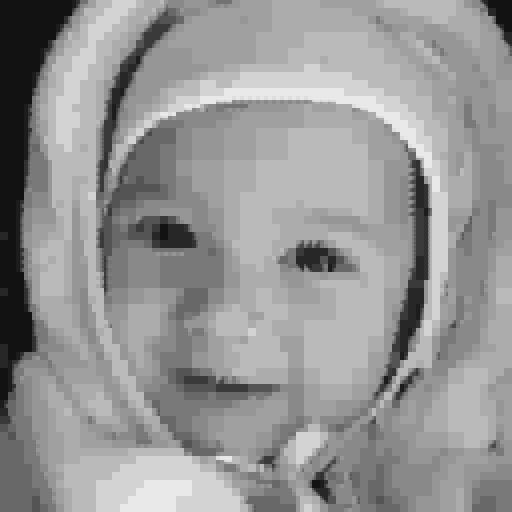} &
\includegraphics[width=2in]{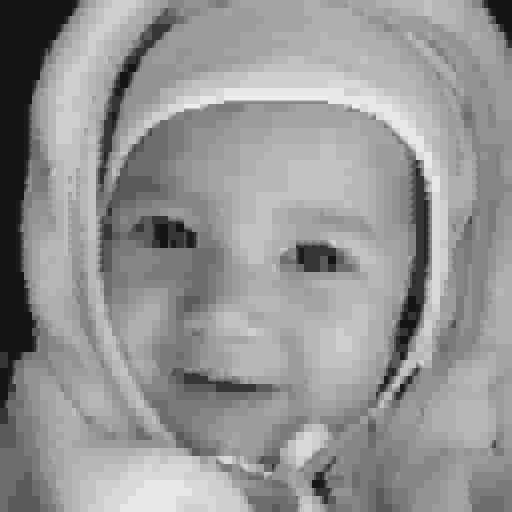} 
\end{tabular}
 
%\end{center}
\caption{\textbf{V}-variable approximations of   Image1 with storage of size 1000 B (PSNR 24.0), 1500 B (PSNR 25.5), 2000 B (PSNR 26.1), and 2500 B (PSNR 26.7).  
%The original picture is stored with $512 \cdot 512=262144$B (since each pix%el is assigned a number in  $\{0,1, \ldots, 255\}$ 
%and thus requires one byte of storage.
%We can thus find a 256-variable approximating image with PSNR slighlty abov%e PSNR 25 at compression ratio 150:1 and we obtain a PSNR around 26.5 at co%mpression ratio 100:1.
}
 \label{fig:enya} 
\end{figure}

  \begin{figure}[H] 
\begin{center}
\begin{tabular}{cc}
\includegraphics[width=2in]{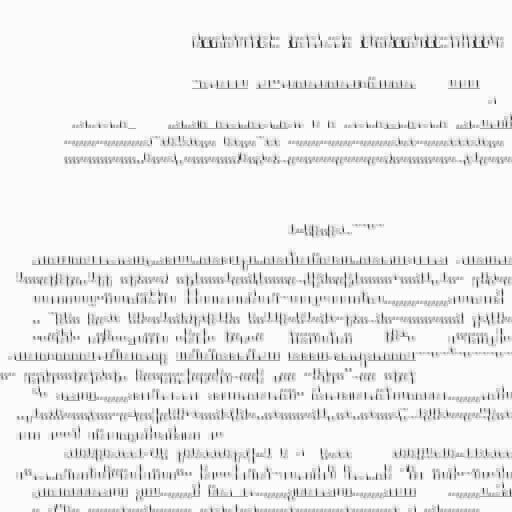} &
\includegraphics[width=2in]{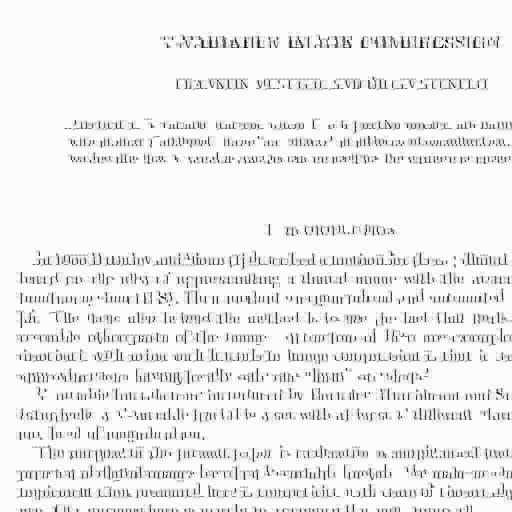} \\ 
\includegraphics[width=2in]{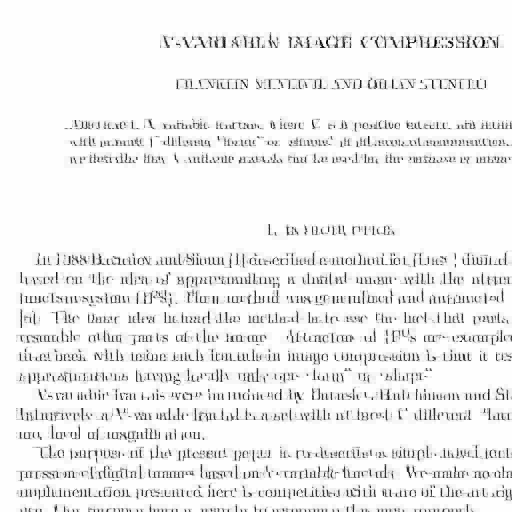} &
    \includegraphics[width=2in]{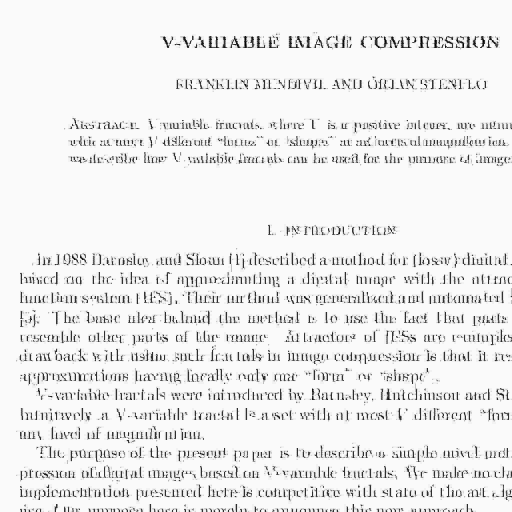} 
\end{tabular}\\

\caption{\textbf{V}-variable approximations of   Image2 with storage of size 1500 B (PSNR 16.7), 3000 B (PSNR 18.3), 4500 B (PSNR 19.4), 6000 B (PSNR 20.4).
  %3000 B (PSNR 17.5), and 4000 B (PSNR 17.9), 5000 B (PSNR 18.1), 6000 B (PSNR 18.1).  
}
\end{center}
 %\begin{tabular}{c}
 % \includegraphics[width=6in]{Scatter256topPaperPNG}
 % \end{tabular}
\end{figure}

\begin{figure}[H] %[hp]   
%\begin{center}
\begin{tabular}{cc}
\includegraphics[width=2in]{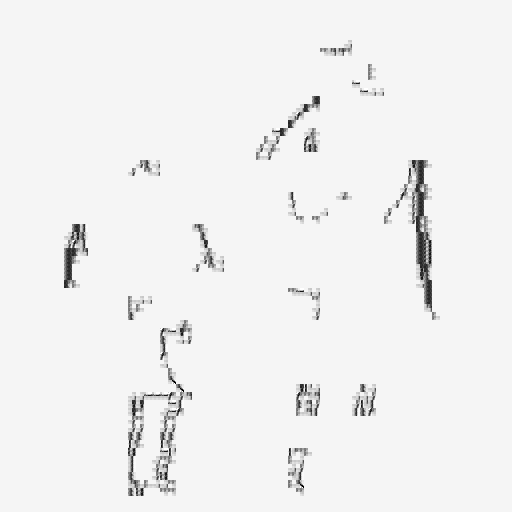} &
\includegraphics[width=2in]{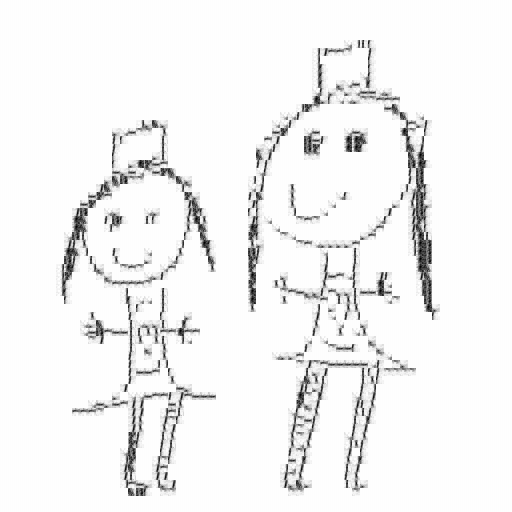} \\
\includegraphics[width=2in]{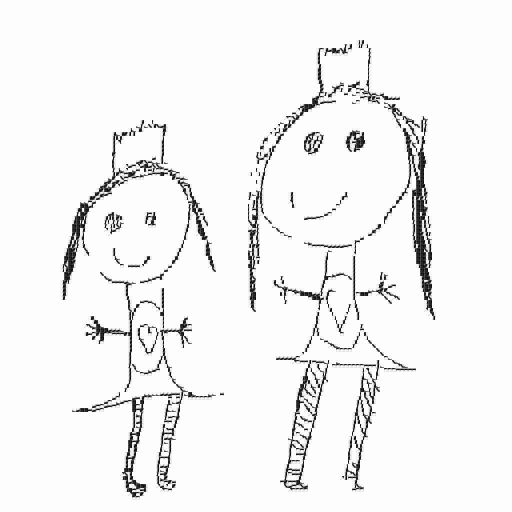} &
\includegraphics[width=2in]{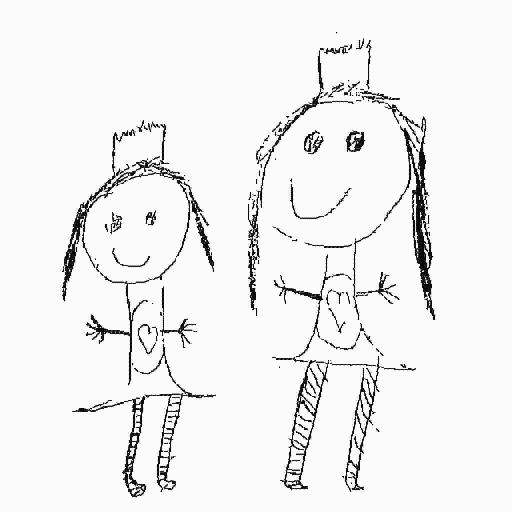} \\
\end{tabular}
\caption{
  \textbf{V}-variable approximations of   Image3 with storage of size 500 B (PSNR 14.6), 1500 B (PSNR 16.5), 2500 B (PSNR 18.4), and 3500 B (PSNR 19.9).  
The approximating images are rather sharp despite having small PSNR values reflecting the fact that the original image is coloured with two extreme colours and any misprediction at colour boundaries gives a high negative influence on the PSNR.}
 \label{fig:approxfigurs}

%\end{center}
\end{figure}

%\newpage

\section{Comparison with other image compression  methods} \label{sec3}

In \cite{MendivilStenflo15} we discussed compression of the test image
here denoted by ``Image1'' with standard fractal block compression.
%Standard fractal block compression with ``small blocks'' of size $16 \times 16$% gives a storage size of 2688 B with PSNR 25.84 for  Image 1 (See \cite{Mendivi%lStenflo15} for details).
%An interesting fact (marked in red in the table below) is that we can obtain the same values for $\mathbf{V}$-variable compression for a certain sequence of% maximally allowed types for the levels.
$\mathbf{V}$-variable compression is slightly better than Fractal compression on Image1 at compression ratios around 100:1 (in the sense of smaller storage size and higher PSNR),
% , see Figure \ref{fig:enya})
 and  much better  on Image2 and Image3. 
%Fractal compression enjoys the property of resolution independence.
%We can also get this with $\mathbf{V}$-variable compression by periodically rep% eating the code.
 Both Fractal compression and $\mathbf{V}$-variable compression are slow methods with respect to compression time, but decompression is fast.

Fourier and wavelet based methods like
JPEG and JP2 works well on photographic images, like Image1, but are not recommended for line art, like Image2 and Image3, where a specialized format like
DjVu works much better.
JPEG gives roughly the same image quality as $\mathbf{V}$-variable image compression at compression ratios around 100:1 for Image1  and better image quality on Image1 for larger file sizes. It is however worse at all compression levels for Image2 and Image3.
This comparison is based on files generated by the free and open source software ImageMagic where we saved the image in JPEG at different quality levels and smaller filesizes than 1.8 KB are then not possible to achieve.
One advantage with $\mathbf{V}$-variable compression compared to JPEG is that image quality is not affected by repeated compression.
One disadvantage with $\mathbf{V}$-variable compression compared to JPEG is that compression time is slower.

JPEG 2000 gives (based on files generated in ImageMagic) better image quality than $\mathbf{V}$-variable image compression at all compression ratios for Image1 and roughly the same image quality for Image2 at high compression ratios. 
The image quality is  worse than $\mathbf{V}$-variable image compression
for Image2 for larger file sizes and worse than $\mathbf{V}$-variable image compression for Image3 
at ``all'' compression ratios.

DjVu is a file format mainly designed for scanned documents that, like $\mathbf{V}$-variable compression,  stores
images containing a combination of text and line drawings, like Image2 and Image3, well.

It seems like $\mathbf{V}$-variable compression could be a competitive format  for intermediate images like digital photos containing line art.

\section{Generalisations} 
\label{general}

\subsection{Non-squared images} \label{nsq}
The general case of a non-squared $j \times k$ image
  can  be treated by  e.g.\ letting $m$ be the smallest integer
such that both $j \leq 2^m$ and $k \leq 2^m$, and consider 
a square image of size $2^m \times 2^m$ where the pixelvalue of a given pixel 
$(a,b)$ in the squared image is given by the pixelvalue of the pixel
 closest to $(j\cdot a/2^m,k \cdot b/2^m)$ in the original image.
We then encode the new squared image and transform the resulting image back in the end.
% Alternatively we can just extend the given image to a   $2^m \times 2^m$ image% , where pixels $(a,b)$ with $a >j$ or $b>k$ are assigned some fixed pixelvalue. 
% We then encode the new squared image and ignore the part outside the $j \times% k$  in the end.

%\subsection{Hybrid methods} 
%Combining $V$-variable and wavelet techniques seems to be efficient tool in compr%essing synthetic images with high compression ratios.
%Another possibility is a hybrid between our $V$-variable approach and trellis-cod%ed quantization. % \cite{SA}.

\subsection{Clustering methods} \label{clusterm}
The efficiency of our algorithm depends crucially on the clustering method we use. 
For simplicity in our implementation above %, in Examples \ref{ex1} and \ref{ex2}
 we  used Matlab's built-in command
{\tt kmeans} for this step. 

  There are many approaches to clustering
and it is an interesting future problem to explore these in order to find
better $\mathbf{V}$-variable approximations than the ones provided by our simple implementation.   In particular, the methods used in vector quantization for constructing the codebook could be explored.
We can also introduce parameters specifying transformations of image pieces as a pre-processing step before clustering.
Matlab also supports ``hierarchical clustering'' which could have been another simple pre-processing alternative.

\subsection{Generalised image pieces}
Our definitions of image pieces, and $\mathbf{V}$-variable images can be generalised in various ways.
We may define the image pieces as arising from some arbitrary given sequence of refining partitions of a given digital image.

% Suppose the partition defining the $n+1$
%th level image pieces, in  a given $j \times k$ pixels image, is obtained from %the $n$th level image pieces by dividing each partition element into $a_{n+1}$ %pieces, $0 \leq n \leq m-1$, and 
% suppose $a_1 a_2 \cdots a_m=j \cdot k$. We may then modify our algorithm by ch%anging $\mathbf{V}$ to $V_n$, and change 
%       step 5 in the algorithm to ``$5.1:\  n=n+1$. $5.2\ $
%Divide each image representative image into $a_{n}$ image pieces and consider t%he $a_{n} V_{n  }$ image pieces created''.

%\begin{Exe}  Consider a  $ 2^m  \times 2^m$ digital image, for some $m$. 
%The algorithm is formulated for the special case when $a_i=4$, and $V_i=V$, for% all $1 \leq i \leq m$ for convenience.
%\end{Exe}

%\begin{Exe}  Consider a  $ 3 \cdot 2^m  \times 2^m$ digital image, for some $m$. 
%A convenient choice for a sequence of refining partitions, is then
% $a_1=3$, and $a_i=4$, $i=2,...,m+1$.             
%\end{Exe}

%\begin{Exe} Consider a  $ 2^m  \times 3^m$ digital image, for some $m$.
%  The choice $a_i=6$, $i=1,...,m$, corresponds to constructing the image pieces% by dividing the image pieces on the present level into $6$ pieces in order to %construct the next level.
%  \end{Exe}

Iterated function systems can be used to recursively define a sequence of
refining partitions.  Thus the IFS machinery gives us great flexibility in
designing recursive partitions and can be used in our scheme.  Any periodic
self-affine tiling is such an example.\\

\subsection{Reducing the colour space}
Reducing the colour space corresponds to chosing the tuple
$\mathbf{V}=(V_1,\ldots,V_m)$ with
$V_m$ small. This corresponds to using a colour palette with $V_m$ different predetermined colours.\\

\noindent
\textbf{Acknowledgment:} We are grateful to  Tilo Wiklund for generous assistance and helpful discussions.
Franklin Mendivil was partially supported by NSERC (2019-05237).  

% other than the IFS in Example \ref{ex:3} as long as its attractor forms a tili%ng of the unit square,

%We can generalize the concept of ``$\mathbf{V}$-variable digital images'' with respect t%o a given sequence of refining 
%partitions generated  by a family of IFSs or by other means. 

% by introducing the following definitions:

%We first define what we mean by a  sequence of refining partitions of an image:
%Let $a(1)$, $a(2)$,$\ldots$,$a(m)=j \cdot k$, be a given sequence of stricly in%creasing positive integers ending with the total number of pixels in the image,% i.e.\ $$ 1 < a(1) < a(2)< ....<a(m)=j\cdot k. $$
%where $m$ is the integer where all image parts are one pixel images.
% (and thus $a(m)=j \cdot k$). 
% For each fixed $n\geq 1$, we divide the   image into $a(n)$ 
%image pieces, in such a way  that each pixel in the image belongs to exactly
% one  image piece. 
%We call these image pieces the image pieces of level $n$.
%Further,  the image pieces are divided in such a way  that each image piece of %level $n+1$  is an image piece of exactly one image piece of level $n$, making %the partition of level $n+1$ into a refinement of the partition of level $n$.

%The $V$-variable image compression algorithm can be generalised in a variety of w% ays.
% We may e.g.\ let the number of distinct image pieces vary from level to level,% use IFSs other than the IFS in Example \ref{ex:3} as long as its attractor for%ms a tiling of the unit square,

\end{document}